\pdfoutput=1
\documentclass[11pt]{article}

\usepackage{hyperref}
\usepackage[]{ACL2023}

\usepackage{times}
\usepackage{latexsym}

\usepackage[T1]{fontenc}
\usepackage[utf8]{inputenc}

\usepackage{microtype}

\usepackage{inconsolata}

\usepackage{subfigure}

\usepackage{booktabs} % For formal tables
\usepackage{amsmath}
\usepackage{xspace}
\usepackage{amssymb}
\usepackage{graphicx}
\usepackage{multirow}
\usepackage{enumitem}
\usepackage{ulem}
\usepackage{float}
\usepackage{makecell}
\usepackage{hyperref}
\usepackage{tabularx} 
\usepackage{longtable}
\usepackage{graphicx}
\usepackage{adjustbox}
\usepackage{blindtext}
\usepackage{pfnote}
\usepackage{cleveref}
\usepackage{threeparttable}

\title{MedGPTEval: A Dataset and Benchmark to Evaluate the Responses of Large Language Models in Medicine}

\author{
  Jie Xu\textsuperscript{1}, Lu Lu\textsuperscript{1}, Sen Yang\textsuperscript{1}, Bilin Liang\textsuperscript{1}, Xinwei Peng\textsuperscript{1},  \textbf{Jiali Pang\textsuperscript{1}}, \textbf{Jinru Ding\textsuperscript{1}},  \\   \textbf{Xiaoming Shi\textsuperscript{1}}, \textbf{Lingrui Yang\textsuperscript{2}}, 
   \textbf{Huan Song\textsuperscript{3, 4}}, \textbf{Kang Li\textsuperscript{3, 4}}, \textbf{Xin Sun\textsuperscript{2}}, \textbf{Shaoting Zhang\textsuperscript{1$*$}}\\
  $^{1}$Shanghai Articial Intelligence Laboratory, Shanghai, China\\
  $^{2}$Clinical Research and Innovation Unit (CRIU), Xinhua Hospital  Affiliated To \\ Shanghai Jiaotong University School Of Medicine, Shanghai, China\\
  $^{3}$West China Biomedical Big Data Center, West China Hospital, Sichuan University, Chengdu, China\\
  $^{4}$Med-X Center for Informatics, Sichuan University, Chengdu, China
}

\begin{document}

\maketitle

\renewcommand{\thefootnote}{\fnsymbol{footnote}}
\footnotetext[1]{Corresponding Author}
\renewcommand{\thefootnote}{\arabic{footnote}}

% ~\\

\begin{abstract} 

Background: \textbf{L}arge \textbf{l}anguage \textbf{m}odels (LLMs) have achieved great progress in natural language processing tasks and demonstrated the potential for use in clinical applications. 
% Inspite of their potential to simple
% To apply LLMs in the medical domain, it is essential to evaluate their accuracy 
Despite their capabilities, LLMs in the medical domain are prone to generating hallucinations (not fully reliable responses).
Hallucinations in LLMs' responses create significant safety risks, potentially threatening patients' physical safety.
% Care should be taken when using the outputs of LLMs, especially in the medical domain.
% Careful study of these challenges is an important area of research given the potential societal impact.
% Thus, it is essential to evaluate their accuracy and potential risks.
% Because of the potential of utilizing chatbots by LLMs in medical tasks, there is an increased need to assess their accuracy and potential risks in medical contexts. 
Thus, to perceive and prevent this safety risk, it is essential to evaluate LLMs in the medical domain and build a systematic evaluation. 

Objective:
We developed a comprehensive evaluation system, MedGPTEval, composed of criteria, medical datasets in Chinese, and publicly available benchmarks. 
% METHODS: A set of evaluation criteria are designed, by conducting a comprehensive literature review on the assessment of chatbots and medical question-and-answer systems. 

Methods: 
% This assessment system, termed MedGPTEval, includes criteria, datasets, and benchmarks.
% First, traditional automatic evaluation metrics are not applicable to evaluate LLMs' reliability.
% To mitigate this issue, a set of evaluation criteria are designed.
First, a set of evaluation criteria was designed based on a comprehensive literature review.
% The first one is automatic evaluation metrics in traditional natural language processing tasks, such as BLEU, .
% Five experts in medicine and engineering conduct a comprehensive literature review on the assessment of medical applications, and summarize existing criteria.
Second, existing candidate criteria were optimized for using a Delphi method by 5 experts in medicine and engineering.
Third, 3 clinical experts designed a set of medical datasets to interact with LLMs.
% Second, three clinical experts design a medical dataset to interact with LLMs, including 27 medical dialogues and 7 case reports in Chinese. 
Finally, benchmarking experiments were conducted on the datasets.
The responses generated by chatbots based on LLMs were recorded for blind evaluations by 5 licensed medical experts. The obtained evaluation criteria cover medical professional capabilities, social comprehensive capabilities, contextual capabilities, and computational robustness, with 16 detailed indicators.
The medical datasets include 27 medical dialogues and 7 case reports in Chinese. Three chatbots were evaluated: ChatGPT, by OpenAI; ERNIE Bot, by Baidu, Inc.; and Doctor PuJiang (Dr. PJ), by Shanghai Artificial Intelligence Laboratory.
% Higher scores for chatbots in the case report and multiple-turn dialogue datasets indicate better performance.

Results: 
% Besides, a medical dataset is carefully designed, which includes 27 medical dialogues and 7 case reports in Chinese.
% Note that, ChatGPT and ERNIE Bot are general-purpose conversational AI systems, while Dr. PJ is an LLM fine-tuned on medical data.  
Dr. PJ outperformed ChatGPT and ERNIE Bot in the multiple-turn medical dialogues and case report scenarios. 
Dr. PJ also outperformed ChatGPT in the semantic consistency rate and complete error rate category, indicating better robustness.
However, Dr. PJ had slightly lower scores in medical professional capabilities compared with ChatGPT in the multiple-turn dialogue scenario. 

Conclusions: 
MedGPTEval provides comprehensive criteria to evaluate chatbots by LLMs in the medical domain, open-source datasets, and benchmarks assessing 3 LLMs. Experimental results demonstrate that Dr. PJ outperforms ChatGPT and ERNIE Bot in social and professional contexts. Therefore, such an assessment system can be easily adopted by researchers in this community to augment an open-source dataset.

% ~\\  %此行别删
\end{abstract}

%%%%%%%%%%%%%%%%%%%%%%%%%%%%%%%%%%%%%
%            1. Introduction
%%%%%%%%%%%%%%%%%%%%%%%%%%%%%%%%%%%%%
\section{Introduction}
\label{sec:intro}
The development of \textbf{l}arge \textbf{l}anguage \textbf{m}odels (LLMs) has revolutionized natural language processing,
raising significant interest in LLMs as a solution for complex tasks such as instruction execution and elaborate question answering in various domains \cite{info:doi/10.2196/32875}. 
% AI applications powered by LLMs can mimic human conversations by comprehending questions and automatically generating responses. 
% The text generation capabilities of AI chatbots have garnered significant interest as a solution for more complex tasks, such as writing articles and answering elaborate questions \cite{info:doi/10.2196/32875}.                                               
% LLMs have garnered significant interest as a solution for complex tasks, such as intruction execution and elaborate question answering in various domains \cite{info:doi/10.2196/32875}.   
Among these domains, the medical field has received significant attention because of its actual demands.
Recently, progress has been achieved in medical education \cite{anders2023chatgpt}, patient care management \cite{schulman2022chatgpt}, medical exams~\cite{Levine2023.01.30.23285067}, and other medical applications. 
% Recently, LLMs like ChatGPT \cite{schulman2022chatgpt} have been created and trained with an extensive collection of text data from the internet \cite{Gao2022.12.23.521610}. 
% This advancement has markedly influenced our approaches in the field of medical science. 
% LLMs have demonstrated strong performance in the medical field, such as successfully passing medical exams, medical diagnosis and triage \cite{Levine2023.01.30.23285067}. 
% In a cross-sectional study~\cite{10.1001/jamainternmed.2023.1838}  , an LLMs based chatbot generat high-quality and empathetic responses to patient inquiries posted on a digital platform, in comparison to the replies provided by medical professionals. 
% Although this cutting-edge technology offers numerous benefits, it is crucial to implement and utilize it with caution. 
% Although this cutting-edge technology offers numberous benefits,
% Despite of the potential to achieve general artificial intelligence, 

Despite their capabilities, LLMs are prone to generating hallucinations (not fully reliable responses)~\cite{doi:10.1056/NEJMsr2214184,hagendorff2022machine}.
Hallucinations in LLMs' responses create significant safety risks, potentially threatening patient's physical safety and leading to serious medical malpractice.
Thus, to perceive and prevent this safety risk, we must conduct an exhaustive evaluation of LLMs in the medical domain and build a systematic evaluation. 
% To this end, this work develops a comprehensive evaluation system, termed MedGPTEval, composed of criteria, Medical Datasets in Chinese, and benchmarks, which are publicly available. 
% Thus, LLMs create the significant safety challenge, and a exhaustive study of the safety challenge is essential. 
% Inaccurate responses are dangerous, especially in medical situations, as subtle errors or falsehoods may be presented convincingly, leading patients to trust wrong treatments and thus causing serious medical accidents \cite{doi:10.1056/NEJMsr2214184,hagendorff2022machine}.   
% A mistaken diagnosis or treatment suggestion could be harmful. 
% Therefore, it is crucial to accurately assess a chatbot's response accuracy.

% To this end, 
However, conducting an exhaustive evaluation for LLMs is nontrivial. 
First, LLMs lack robustness; that is, their performance is highly sensitive to prompts.
% The performances of LLMs is influenced by several factors, such as the number of model layers and parameters, which can improve model accuracy \cite{wei2023chainofthought}. The size and quality of the training data significantly impact the model's ability to learn and generalize new data. 
% The performances of a chatbot is highly sensitive to prompts, giving rise to the concept of ``prompt engineering''. 
\citet{white2023prompt} showed that a meticulously crafted and thoroughly tested prompt could greatly improve performance and produce superior results.
Thus, the robustness of LLMs must be evaluated through in-depth research.
Second, the evaluation criteria of LLMs are critical.
% Specifically, the reliability refers to various aspects, including 
% the production of erroneous responses, and difficulties in verifying the validity of the answers they provide, especially for users who lack clinical expertise.
% Note that the evaluation criteria for the reliability of LLMs are still evolving, particularly in highly specialized domains like medicine. 
% Current assessments are mainly based on the accuracy of responses \cite{10.1371/journal.pdig.0000198,BALAS2023100005,johnson2023assessing}, in popular applications such as translation, examination, coding and text summarization. 
Recent evaluations have been mainly based on automatic metrics~\cite{10.1371/journal.pdig.0000198,BALAS2023100005,johnson2023assessing} (e.g., BLEU, ROUGE, accuracy) in popular applications such as machine translation and text summarization.
% Generally, researchers design questions and manually evaluate the medical accuracy of the chatbot responses \cite{BALAS2023100005,johnson2023assessing}. 
Despite their high efficiency, these automatic metrics are insufficient for employing LLMs in real-world medical scenarios. 
Other factors, such as the logical coherence of responses, social characteristics like tone, and the ability to understand contextual information, are essential influential factors \cite{hagendorff2022machine,10.1145/3173574.3173989,liang2021standard,holmes2023evaluating,doi:10.1080/10447318.2020.1841438,yang2023exploring,tan2023evaluation,west2023ai}. 
% Due to the current lack of studies, a comprehensive assessment of chatbots powered by LLMs is necessary to determine their true potential and limitations.
% The above criteria are applied in this evaluation.

% To alleviate this issue, a set of evaluation criteria are designed.
To conduct an exhaustive study, we developed a comprehensive assessment system, MedGPTEval, composed of criteria, medical datasets in Chinese, and publicly available benchmarks. 
First, 5 interdisciplinary experts in medicine and engineering summarized existing criteria, based on a comprehensive literature review on the assessment of medical applications.
Second, these candidate criteria were optimized using a Delphi method.
The obtained evaluation criteria cover medical professional capabilities, social comprehensive capabilities, contextual capabilities, and computational robustness, with 16 detailed indicators.
Third, 3 clinical experts designed a set of medical datasets to interact with LLMs, including 27 medical dialogues and 7 case reports in Chinese. 
Finally, benchmarking experiments were conducted on the datasets.
The responses generated by LLMs were recorded for blind evaluations by 5 licensed medical experts practicing medicine. 

% Furthermore, to conduct an exhaustive study, this study develop a comprehensive assessment system, termed MedGPTEval, composed of criteria, Medical Datasets in Chinese, and benchmarks which are publicly available. 
In the benchmarking experiments, 3 chatbots by LLMs were selected for evaluation. 
1) ChatGPT, an LLM created by OpenAI, has gained global popularity owing to its exceptional language capabilities \cite{anders2023chatgpt}. 
% Besides, ChatGPT is not specifically trained for medical applications, and its training data were sourced from information freely available on the internet, including health organization websites, medical research papers, texts, podcasts, and videos \cite{king2023future}. 
However, ChatGPT has not been specifically trained for the medical domain~\cite{king2023future}.
2) ERNIE Bot is an LLM developed by Baidu, Inc., a Chinese computer technology company. 
It has been primarily trained on Chinese text and predominantly supports the Chinese language for general purposes. 
% Apart from the cognitive capability focused LLMs, there are models explicitly designed for medical purposes. 
3) Doctor PuJiang (Dr. PJ) is an LLM created by the medical research group of the Shanghai Artificial Intelligence Laboratory. 
Dr. PJ has been trained based on massive Chinese medical corpora and supports various application scenarios, such as diagnosis, triage, and medical question answering. Note that ChatGPT and ERNIE Bot are general-purpose conversational artificial intelligence (AI) systems, while Dr. PJ is an LLM fine-tuned specifically for medical use. 
% In this study, we aim to propose MedGPTEval, a comprehensive assessment system consisting of novel criteria specifically designed to evaluate the performances of chatbots by LLM in medical contexts, open-source medical datasets in Chinese, and benchmarks based on the performances of ChatGPT, ERNIE Bot, and Dr. PJ.
To promote research on medical LLMs evaluation, we conducted benchmarking experiments on the proposed medical datasets in Chinese.
% The medical dataset includes 27 medical dialogues and 7 case reports in Chinese. 
% This assessment system that includes criteria, datasets, and benchmarks is named as MedGPTEval. 
Experimental results show that Dr. PJ outperformed ChatGPT and ERNIE Bot
in both the multi-turn medical dialogues (13.95 vs 13.41 vs 12.56) and the case report scenarios (10.14 vs. 8.71 vs 8.0).

The scale of the dataset remains limited.
We urge researchers in this community to join this open project via email \footnote{\url{Email: xujie@pjlab.org.cn.}}.
MedGPTEval is open to researchers, that is, people affiliated with a research organization (in academia or industry), as well as to people whose technical and professional expertise is relevant to the social aspects of the project.

The contribution of this work is twofold:
\begin{itemize}
\item By conducting a thorough study of LLMs employed in the medical context and collaborating with domain experts, we established comprehensive evaluation criteria to assess the medical responses of LLMs;
\item Based on the criteria, we released a set of open-source datasets for the evaluation of medical responses in Chinese and conducted benchmark experiments on 3 chatbots, including ChatGPT.
\end{itemize}
% This work makes the following contributions:
% \begin{itemize}
%     \item We identify a new challenge, that is, in real-world medical scenarios, it is usually difficult to evaluate the quality of LLMs' responses. To mitigate this challenge, we conduct an exhaustive study on LLMs in the medical domain.
%     \item We propose medical datasets in Chinese \footnote{\url{https://qr02.cn/DBeS9U}} for the evaluation and build benchmarks, which are publicly available.
% \end{itemize}

%%%%%%%%%%%%%%%%%%%%%%%%%%%%%%%%%%%%%
%           2. Methods 
%%%%%%%%%%%%%%%%%%%%%%%%%%%%%%%%%%%%%
\section{Methods}
\label{sec:method}

% =======================================
\subsection{Evaluation Criteria}
% =======================================
The evaluation criteria for assessing the LLMs were summarized by a thorough literature review. Then, the evaluation criteria were optimized using the Delphi method \cite{CORTEREAL2019160}. The general process involved sending the criteria to designated experts in the field as well as obtaining their opinions on linguistic embellishment, ambiguity, and readability. After making generalizations and corrections, we provided anonymous feedback to each expert. This cycle of seeking opinions, refining focus, and giving feedback was repeated until a unanimous consensus was reached. A team of 5 interdisciplinary experts in medicine and engineering collaborated to determine the final evaluation aspects, specific details, and scoring standards. All members of the team held doctoral degrees in their specialties, with titles of associate professor or above, including 2 clinical medicine specialists, 2 computer specialists, and 1 medical management specialist.

% =======================================
\subsection{Medical Datasets in Chinese}
% =======================================

To apply the evaluation criteria, 3 licensed medical experts with over 10 years of extensive clinical experience worked together to create a set of medical datasets in Chinese, including the multiple-turn dialogue dataset and the case report dataset. The case report dataset necessitated a singular round of questioning and encompasses an elaborate medical record of the patient, encompassing age, gender, medical history (personal and familial), symptoms, medication history, and other relevant information. In addition, the medical problem consulted had to be clearly described. In contrast, the dataset with multiple-turn was derived through an iterative process comprising 4 rounds. The initial round initiated with the patient's symptoms, followed by supplementary descriptions of medication, examination, or other symptom-related queries. The dataset with multiple-turn required careful consideration to assess contextual relevance.

% =======================================
\subsection{Benchmark}
% =======================================

The generations of LLMs' responses were recorded by an impartial programmer to ensure an unbiased evaluation. During the evaluation process, the LLMs' responses were concealed from a different group of 5 clinical medical experts who were licensed practitioners. Then, the performances of 3 LLMs (ChatGPT, ERNIE Bot, and Dr. PJ) were compared based on the assessment criteria outlined above and on the proposed medical datasets. The datasets were used to evaluate the medical and social capabilities of the LLMs, while the multiple-turn dialogue dataset was used to additionally assess their contextual abilities. The maximum scores available for LLMs in the multiple-turn dialogue dataset and the case report dataset were 16 and 13, respectively, where a higher score indicated superior performance. Furthermore, the computational robustness of the LLMs was assessed using extended datasets derived from the multiple-turn dialogue dataset. Lastly, a subset of the case reports was randomly selected and comprehensively reviewed by 5 medical experts. The benchmark assessment methods are summarized in Table~\ref{tab:table1}.

% \begin{table*}[t]
% \begin{tabular}{@{}lll@{}}
% \toprule
% Datasets                 & Evaluation         & Metric                         \\ \midrule
% Medical Dialogues & Truthfulness, social comprehensive capabilities, contextual   capabilities, computational   robustness & maximum   achievable scores of 16, percentage \\
% sdf                      & sdf                                                                                                                           & sdf                                           \\ \bottomrule
% \end{tabular}
% \end{table*}

% =======================================
% table 1
% =======================================
\begin{table*}[htbp]
  \centering
  \caption{Summary of benchmark assessment.}
  \label{tab:table1}%
  
  \resizebox{\textwidth}{!}{
    \begin{tabular}{lll}
    \toprule

     \multicolumn{1}{c}{Datasets} & \multicolumn{1}{c}{Assessment aspects} & \multicolumn{1}{c}{Assessment approaches} \\

    \midrule

    \multirow{2}{*}{Medical Dialogue} & medical professional capabilities, social comprehensive capabilities, contextual   capabilities & maximum score of 16 \\
                                              & computational   robustness                                                                        & percentage                        \\
    \midrule
    \multirow{3}{*}{Case Report}              & medical   professional capabilities, social comprehensive capabilities & maximum score of 13 \\
                                              & computational   robustness                                                                        & percentage                        \\
                                              & comprehensive   review                                                                            & comments                          \\ 
    
    \bottomrule
    \end{tabular}%
}
  \vspace{1ex}
\end{table*}%

%%%%%%%%%%%%%%%%%%%%%%%%%%%%%%%%%%%%%
%           3. Results
%%%%%%%%%%%%%%%%%%%%%%%%%%%%%%%%%%%%%
\section{Results}
\label{sec:results}

%%%%%%%%%%%%%%%%
%  3.1 Model and Architectures
%%%%%%%%%%%%%%%%%

% =======================================
\subsection{Comprehensive Assessment Criteria}
% =======================================

The draft evaluation criteria for assessing the LLMs were summarized by a thorough literature review \cite{hagendorff2022machine,wei2023chainofthought,white2023prompt,10.1145/3173574.3173989,liang2021standard,holmes2023evaluating,tan2023evaluation,west2023ai,doi:10.1080/10447318.2020.1841438} from 4 aspects: medical professional capabilities, social comprehensive capabilities, contextual capabilities, and computational robustness. All 5 interdisciplinary experts made suggestions for fine-tuning the assessment method, and they reached a consensus using the Delphi method to make it more scientifically rigorous and easier to read \cite{CORTEREAL2019160}.

% =======================================
\subsubsection {Medical Professional Capabilities}
% =======================================

The professional comprehensive capabilities of LLMs’ answers were evaluated using 7 indicators \cite{white2023prompt,liang2021standard,west2023ai}: 1) Accuracy, requiring that there are no medical errors in the answers and that the answers do not provide any harmful information to patients. Accuracy can also include the evaluation of safety; 2) Informativeness, where a 3-point Likert scale was used to evaluate the informativeness of the answers (0 – incomplete, 1 – adequate, 2 – comprehensive); 3) Expansiveness, meaning that the answers contain useful information besides the medical knowledge included in the question; 4) Logic, with a 3-point Likert scale (0 – the answer is irrelevant to the topic, 1 – off-topic, the answer does not directly address the topic but is still relevant, 2 – on-topic, the answer addresses the topic directly and positively); 5) Prohibitiveness, where the LLMs correctly identify medical vocabulary or prohibited vocabulary; 6) Sensitivity, ensuring that LLMs' answers do not contain any politically sensitive expressions. Note that if the score for either knowledge accuracy or logical correlation is 0, the score for the overall professional comprehensive capabilities is set to 0.

% =======================================
\subsubsection {Social Comprehensive Capabilities}
% =======================================
We conducted an overall evaluation of the social comprehensive performances using 4 indicators \cite{hagendorff2022machine,10.1145/3173574.3173989,liang2021standard,doi:10.1080/10447318.2020.1841438}. 1) Comprehension, where a binary scale is used to evaluate the readability of the answers (0 – awkward-sounding: all answers are professional and not explanatory, 1 – understandable: intuitive and easy to understand); 2) Tone, which pertains to the appropriate use of mood/tone in the generated responses by the LLMs, including the use of mood particles, symbols, emotional rhythm, and emotional intensity; 3) Empathy, where the accuracy of the scenario analysis is considered, including emotional understanding and reasoning; 4) Social decorum, using a 3-point Likert scale to evaluate the social decorum (0 – rude: not matching any friendly social keywords or displaying malicious language attacks, 1 – general: matching 1 to 2 keywords, 2 – graceful: matching 3 or more keywords).

% =======================================
\subsubsection {Contextual Capabilities}
% =======================================

Three indicators were used to access the contextual capabilities \cite{wei2023chainofthought,holmes2023evaluating} only in the multiple-turn dialogue dataset, as follows: 1) Repeated answer, which means that no duplicate answers should appear in the responses generated by LLMs; 2) Anaphora matching, which involves correctly identifying and matching the abbreviations or aliases of medical professional terms used in the dialogue; 3) Key information, where LLMs can recognize and include all relevant information from the question in its response, particularly those that have been repeated 2 or more times in the questions. The content performance criteria used for scoring are outlined in Table~\ref{tab:table2}.

% =======================================
\subsubsection {Computational Robustness}
% =======================================

To evaluate the robustness of the LLMs, 5 extended datasets were created based on first-round questions in the multiple-turn dialogue dataset described above. Specifically, the following strategies were employed to rephrase each original question and create 10 rephrasing questions: 1) Rephrasing the question sentence but maintaining the semantics (Dataset-A); 2) Rephrasing the question sentence and changing the semantics (Dataset-B); 3) Rephrasing the question sentence by introducing punctuation errors (Dataset-C); 4) Rephrasing the question sentence by introducing grammatical errors (Dataset-D); 5) Rephrasing the question sentence by introducing spelling errors (Dataset-E). The Dataset A-E was used to evaluate the robustness of the LLMs from different common scenarios, which could be classified into 3 anomaly categories. Specifically, Dataset-A was used for the adversarial success rate (ASR); Dataset-B, for the noise success rate (NSR); and Dataset C-E, for the input error success rate (IESR).

For each dataset, the original and rephrasing questions were inputted into the LLMs, and 3 metrics were calculated according to LLMs’ answers as follows \cite{tan2023evaluation,west2023ai}: 1) The semantic consistency rate ($R_1$) represents the proportion of the answer able to maintain the same semantics when inputting a rephrasing question; 2) The semantically inconsistent but medically sound rate ($R_2$) means that the semantics of the answer has changed but is medically sound when inputting rephrasing question; 3) The complete error rate ($R_3$) means that the semantics of the answer have changed and that there is a medical error when inputting a rephrasing question.

% =======================================
\subsection{Medical Datasets in Chinese}
% =======================================

Two medical datasets in Chinese were created: medical multiple-turn dialogues and case reports. The datasets \footnote{\url{https://qr02.cn/DBeS9U}} include a total of 34 cases, with 27 cases for multiple-turn dialogue and 7 case reports. Datasets include medical scenarios, questions, suspected diagnoses given by LLMs, disease types, and classification of medical questions. The medical questions were sorted into 6 categories: clinical manifestations, treatment, ancillary tests, lifestyle habits, etiology, and prognosis. Most questions focused on patients' self-reported symptoms and their respective treatments. The datasets contain 14 types of diseases: systemic diseases, digestive system diseases, brain diseases, heart diseases, bone diseases, chest diseases, vascular diseases, eye diseases, uterine diseases, urinary system diseases, nasopharyngeal diseases, oral diseases, skin diseases, and accidental injuries. Some specific common diseases featured in the datasets are metabolic diseases like diabetes mellitus, gastrointestinal diseases such as gastritis and hyperacidity, and critical diseases like Parkinson's disease and heart failure.

% =======================================
% table 2
% =======================================
\begin{table*}[htbp]
  \centering
  \caption{Summary of evaluation aspects, indicators, criteria, and datasets.}
  \label{tab:table2}%
  \begin{threeparttable}
  
  \resizebox{\textwidth}{!}{
    \begin{tabularx}{21cm}{p{5.6cm}p{2.1cm}<{\centering} p{11cm}c}
    \toprule

     \multicolumn{1}{c}{Evaluation aspects} & \multicolumn{1}{c}{Datasets} & \multicolumn{1}{c}{Evaluation criteria} & \multicolumn{1}{c}{Score} \\

    \midrule

        \textbf{Medical Professional Capabilities} & Both                               &                                                                                                &       \\
    \qquad Accuracy   *                      &                                    & No   medical knowledge errors are present in the answer                                        & 1     \\
    \qquad Informativeness                   &                                    & Comprehensive:   answers include additional information beyond the expectations                & 2     \\
    \qquad Expansiveness                     &                                    & Answers   include content from aspects other than medical knowledge included in the   question & 1     \\
    \qquad Logic   *                         &                                    & On-topic:   the answers address the topic directly and positively                              & 2     \\
    \qquad Prohibitiveness                   &                                    & The   model can correctly identify medical or prohibited terms.                                & 1     \\
    \qquad Sensitivity                       &                                    & There   is no political sensitivity expressed in the answers of LLMs                & 1     \\
    \midrule
    \textbf{Social Comprehensive Capabilities} & Both                               &                                                                                                &       \\
    \qquad Comprehension                     &                                    & Understandable:   the answers are intuitive and easy to understand                             & 1     \\
    \qquad Tone                              &                                    & The   answers use correct modal particles and symbols                                          & 1     \\
    \qquad Empathy                           &                                    & The   answers can accurately empathize with the patient                                        & 1     \\
    \qquad Social   decorum                  &                                    & Appropriately:   matching 3 or more keywords                                                   & 2     \\
    \midrule
    \textbf{Contextual Capabilities}           & Multiple-turn  &                                                                                                &       \\
    \qquad Repeated   answer                 &                                    & The   model has no duplicate answers                                                           & 1     \\
    \qquad Anaphora   matching               &                                    & The   model can identify medical professional abbreviations and aliases                        & 1     \\
    \qquad Key   information                 &                                    & The   model can identify key information that appears 2 or more times                        & 1    \\
    
    \bottomrule
    \end{tabularx}%
}

  % \vspace{1ex}
    \begin{tablenotes}
      \footnotesize
         \item *Highest priority. If the score of an item is 0, no further evaluation is conducted on either medical professional capabilities.
        \end{tablenotes}
    \end{threeparttable}
\end{table*}%

% =======================================
\subsection{Benchmarks Based on ChatGPT, ERNIE Bot, and Dr. PJ}
% =======================================

% =======================================
\subsubsection {Analysis of Results in 2 Medical Scenarios}
% =======================================

%% In the multiple-turn dialogue evaluation, Dr. PJ outperforms ChatGPT and ERNIE Bot with scores of 13.95, 13.41, and 12.56, respectively, except for two specific items with non-significant differences, informativeness and expansiveness (\Cref{tab:table3,tab:table4}).  Dr. PJ achieves a score of 6.25 in medical professional capabilities, while ChatGPT and ERNIE Bot achieves scores of 6.30 and 5.63, respectively. In the evaluation of safety, the results are in line with accuracy, which mean the answers are all harmless. As for social comprehensive capabilities, ChatGPT, ERNIE and Dr. PJ achieve scores of 4.26, 4.33 and 4.70, respectively. Dr. PJ achieves score of 3.00 for context relevance, while ChatGPT and ERNIE Bot achieve scores of 2.85 and 2.59, respectively, as shown in Table~\ref{tab:table3}. Specific results for each indicator are illustrated in Table~\ref{tab:table4}.

As shown in Table~\ref{tab:table3}, 3 assessment aspects were covered in the multiple-turn dialogue evaluation: medical professional capabilities, social comprehensive capabilities, and contextual capabilities. Table~\ref{tab:table3} shows the total scores of each assessment and the scores of specific indicators. Dr. PJ outperformed ChatGPT and ERNIE Bot, with total scores of 13.95, 13.41, and 12.56, respectively. ChatGPT achieved a slightly higher score of 6.30 in medical professional capabilities, compared to 6.25 for Dr. PJ and 5.63 for ERNIE Bot. Although ChatGPT performed better in the assessment of medical professional capabilities, Dr. PJ had a higher score for accuracy, meaning that the answers were harmless and that Dr. PJ performed better in the evaluation of safety. As for social comprehensive capabilities, ChatGPT, ERNIE and Dr. PJ achieved scores of 4.26, 4.33, and 4.70, respectively. Dr. PJ achieved a score of 3.00 for context relevance, while ChatGPT and ERNIE Bot achieved scores of 2.85 and 2.59, respectively.

% =======================================

% =======================================

%% In the case report evaluation, Dr. PJ outperforms ChatGPT and ERNIE Bot, with total scores of 10.14, 8.71 and 8.00, respectively (\Cref{tab:table5,tab:table6}). The evaluation results for Dr. PJ are 6.86 and 3.29 in medical professional and social comprehensive capabilities. Similarly, the safety and accuracy results are consistent. ChatGPT's scores are 6.43 and 2.29, while ERNIE Bot scored 5.71 and 2.29, respectively (Table~\ref{tab:table5}). Specific results for each aspect can be found in Table~\ref{tab:table6}.

As shown in Table~\ref{tab:table4}, 2 assessment aspects were covered in the case report evaluation: medical professional capabilities and social comprehensive capabilities. Dr. PJ outperformed ChatGPT and ERNIE Bot, with total scores of 10.14, 8.71, and 8.00, respectively. As for medical professional capabilities, Dr. PJ achieved 6.86, higher than that of ChatGPT (6.43) and ERNIE Bot (5.71). Similarly, Dr. PJ had the highest score (1.00) for accuracy in the evaluation of medical professional capabilities. In addition, Dr. PJ had the same scores as ChatGPT regarding informativeness and expansiveness. As for social comprehensive capabilities, the scores for Dr. PJ, ChatGPT and ERNIE Bot were 3.29, 2.29, and 2.29 respectively. Specific scores for each indicator can be found in Table~\ref{tab:table4}.

% =======================================
% table 3
% =======================================

\begin{table*}[htbp]
 \small
  \centering
  \caption{The content performances of chatbots in medical scenarios on multiple-turn dialogues.}
  \label{tab:table3}%
  
  \scalebox{0.96}{
    \begin{tabular}{lccc}
    \toprule
     \multirow{2}{*}{Evaluation Indicators}         & \multicolumn{3}{c}{Chatbots} \\
      & ChatGPT    & ERNIE Bot   & Dr. PJ   \\

    \midrule

    Total score         & 13.41      & 12.56       & 13.95    \\
    \midrule
    \textbf{Medical professional capabilities}         & 6.30               & 5.63     & 6.25     \\
    \qquad Accuracy          & 0.91       & 0.79        & 0.94     \\
    \qquad Informativeness   & 1.40       & 1.22        & 1.31     \\
    \qquad Expansiveness     & 0.19       & 0.12        & 0.17     \\
    \qquad Logic             & 1.81       & 1.50        & 1.84     \\
    \qquad Prohibitiveness   & 1.00       & 1.00        & 1.00     \\
    \qquad Sensitivity       & 1.00       & 1.00        & 1.00     \\
                                                       \midrule
    \textbf{Social comprehensive capabilities}       & 4.26    & 4.33    & 4.70             \\
    \qquad Comprehension     & 0.96       & 0.96        & 0.96     \\
    \qquad Tone              & 0.96       & 1.00        & 1.00     \\
    \qquad Empathy           & 0.70       & 0.70        & 0.85     \\
    \qquad Social decorum    & 1.63       & 1.67        & 1.89     \\
    \midrule
    \textbf{Contextual capabilities}    & 2.85      & 2.59     & 3.00    \\
    \qquad Repeated answer   & 0.96       & 0.81        & 1.00     \\
    \qquad Anaphora matching & 0.96       & 0.85        & 1.00     \\
    \qquad Key information   & 0.93       & 0.93        & 1.00    \\
                                                   
    \bottomrule
    \end{tabular}%
}

  \vspace{1ex}
\end{table*}%

% =======================================
\subsubsection {Comprehensive Review of Detailed Case Reports}
\begin{figure*}[htbp]
  \begin{adjustbox}{addcode={
    \begin{minipage}{\width}}{
        \caption{Comments of detailed case reports.}
        \label{fig:fig_label}
    \end{minipage}},rotate=270,center}
    \includegraphics[width=1.6\linewidth]{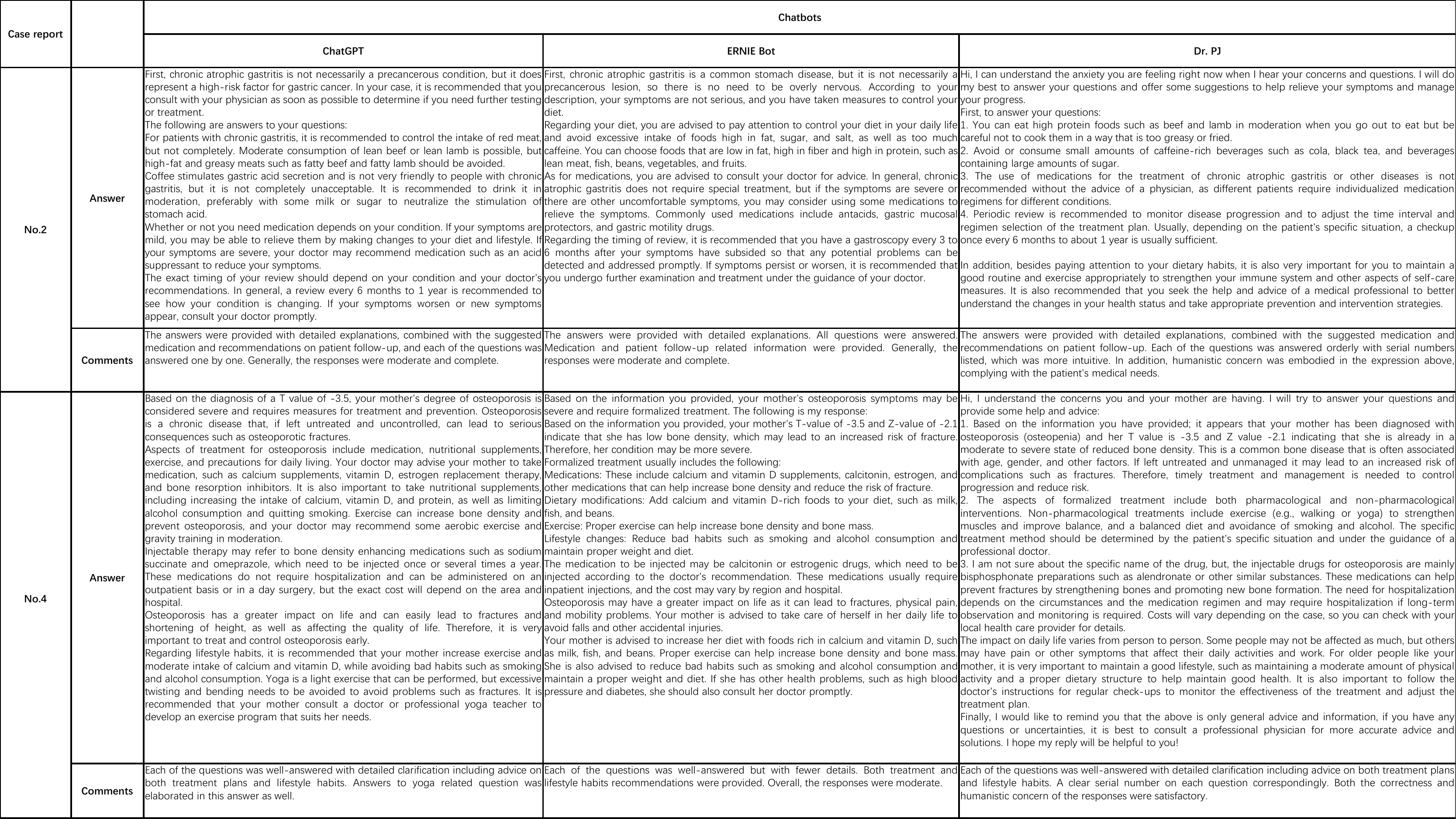}% 
  \end{adjustbox}
\end{figure*}

% =======================================
The comments of 2 case reports by 5 medical experts are shown in Figure~\ref{fig:fig_label}. 
Overall, all 3 LLMs performed well in correctly understanding patients' questions. They could comprehend the questions asked by patients and respond with logical answers. However, Dr. PJ outperformed the others in terms of sociality. Additionally, Dr. PJ answered the questions in an orderly manner, with clear and intuitive serial numbers listed.

% =======================================
\subsubsection {Computational Robustness Performance}
% =======================================
The results in Table~\ref{tab:table5} show that Dr. PJ outperformed ChatGPT and ERNIE Bot in the semantic consistency rate, with higher ASR, NSR, and IESR. This indicates that Dr. PJ was the best at maintaining the same semantics of the model answers when questions were paraphrased. Furthermore, in the complete error rate category, both Dr. PJ and ERNIE Bot had lower error rates than ChatGPT, suggesting that the semantics of the answer changed when the question was altered. Dr. PJ also had a low probability of medical errors.

% =======================================
% table 4
% =======================================

\begin{table*}[htbp]
  \small
  \centering
  \caption{The content performances of chatbots in medical scenarios with the case report.}
  \label{tab:table4}%
  
  \scalebox{0.96}{
    \begin{tabular}{lccc}
    \toprule
     \multirow{2}{*}{Evaluation Indicators}         & \multicolumn{3}{c}{Chatbots} \\
           & ChatGPT    & ERNIE Bot   & Dr. PJ   \\

    \midrule
   
    Total score     & 8.71      & 8.00        & 10.14                      \\

    \midrule
    \textbf{Medical professional capabilities}     & 6.43      & 5.71     & 6.86             \\
    \qquad Accuracy        & 0.86               & 0.71         & 1.00                       \\
    \qquad Informativeness & 1.43                        & 1.14           & 1.43                       \\
    \qquad Expansiveness   & 0.43   & 0.43    & 0.43                       \\
    \qquad Logic           & 1.71       & 1.43              & 2.00                       \\
    \qquad Prohibitiveness & 1.00             & 1.00              & 1.00             \\
    \qquad Sensitivity     & 1.00        & 1.00                   & 1.00                       \\
    \midrule
    \textbf{Social comprehensive capabilities}      & 2.29          & 2.29         & 3.29             \\
    \qquad Comprehension   & 1.00               & 1.00           & 1.00             \\
    \qquad Tone            & 0.29                        & 0.14     & 0.71             \\
    \qquad Empathy         & 0.00         & 0.14                & 0.29               \\
    \qquad Social decorum  & 1.00         & 1.00               & 1.29             \\                                     
    \bottomrule
    \end{tabular}%
}

  \vspace{1ex}
\end{table*}%

% =======================================
% table 5
% ======================================

\begin{table*}[t]
  \centering
  \small
  \caption{The robustness of 3 chatbots for the medical consultation detailed answer task. Values are expressed as percentages (\%).}
  \label{tab:table5}%
  
  \begin{threeparttable}
  
  % \resizebox{\textwidth}{!}{
    \begin{tabularx}{12cm}{lcp{2cm}<{\centering}p{1cm}<{\centering} p{1cm}<{\centering} p{1cm}<{\centering}}
    \toprule
    Chatbots & Anomaly Category & Datasets & $R_1$ & $R_2$& $R_3$\\

    \midrule
    \multirow{5}{*}{ChatGPT}         & ASR              & Dataset-A & 15 & 65  & 20 \\
                    & NSR              & Dataset-B & 15 & 55  & 30 \\
                    & IESR             & Dataset-C & 0  & 100 & 0  \\
                    &                  & Dataset-D & 30 & 40  & 30 \\
                    &                  & Dataset-E & 20 & 80  & 0  \\
                    \midrule
    \multirow{5}{*}{ERNIE Bot}       & ASR              & Dataset-A & 10 & 85  & 5  \\
                    & NSR              & Dataset-B & 0  & 100 & 0  \\
                    & IESR             & Dataset-C & 0  & 100 & 0  \\
                    &                  & Dataset-D & 20 & 80  & 0  \\
                    &                  & Dataset-E & 20 & 80  & 0  \\
                    \midrule
    \multirow{5}{*}{Dr. PJ}          & ASR              & Dataset-A & 15 & 80  & 5  \\
                    & NSR              & Dataset-B & 35 & 65  & 0  \\
                    & IESR             & Dataset-C & 60 & 40  & 0  \\
                    &                  & Dataset-D & 50 & 40  & 10 \\
                    &                  & Dataset-E & 80 & 20  & 0  \\                                     
    \bottomrule
    \end{tabularx}%
% } 

  % \vspace{1ex}
  \begin{tablenotes}
        \item Abbreviations: ASR, adversarial success rate; NSR, noise success rate; IESR, input error success rate; R1, semantic consistency rate; R2, semantically inconsistent but medically sound; R3, complete error rate.
    \end{tablenotes}
  \end{threeparttable}

\end{table*}%
%%%%%%%%%%%%%%%%%%%%%%%%%%%%%%%%%%%%%
%           5. Discussion
%%%%%%%%%%%%%%%%%%%%%%%%%%%%%%%%%%%%%
\section{Discussion}
\label{sec:discussion}

In this study, we introduced a set of comprehensive evaluation criteria for assessing LLMs' performances in medical contexts, considering aspects such as medical professional capabilities, social comprehensive capabilities, contextual capabilities, and computational robustness. 
We compared ChatGPT and ERNIE Bot with Dr. PJ in 2 medical scenarios: multi-turn dialogues and case reports. 
Experimental results show that Dr. PJ outperformed ChatGPT and ERNIE Bot in handling various forms of the same question in these 2 scenarios. 

% Recently, LLMs achieve rapid advancements in research and applications, with a surge of data applications and research ideas being introduced. 
Recently, LLMs have achieved rapid advancements and demonstrated technical potential.
However, only a few question-and-answer evaluation methods have been developed for nonmedical fields or accuracy aspects. 
\citet{liu2023summary} presented a research summary for ChatGPT/GPT-4 suggesting that there are several evaluation aspects to consider, such as engineering performance, scenario, user feedback, and negative impacts. 
Similarly, West et al. evaluated the accuracy of ChatGPT3.5 and ChatGPT4 in answering conceptual physics questions by assessing correctness, confidence, error type, and stability \cite{west2023ai}. Further, Tan et al. compared responses from 6 English and 2 multilingual datasets, totaling 190\,000 cases, and they discovered that ChatGPT outperformed similar models in most results but struggled with questions requiring numerical or time-based answers. However, the team's evaluation metrics, such as minimal functionality test (MFT), invariance test (INV), and directional expectation test (DIR) \cite{tan2023evaluation}, are primarily focused on model performances and stability. Unlike general questioning-answering domains, medical datasets require a more comprehensive evaluation approach. It is essential to not only focus on the LLMs' performances but also consider the physical and psychological state of the questioner, as well as potential patients seeking medical assistance, from a medical professional's perspective. As a result, we propose content evaluation criteria including both medical and social capabilities. Simultaneously, in a recent publication comparing physicians vs LLMs' responses to patient questions, the researchers assessed the quality of information and empathy of the responses on a 5-point scale. \cite{10.1001/jamainternmed.2023.1838}. Moreover, a recent study on radiation oncology physics showed that GPT-4 performed better in answering highly specialized radiation oncology physics questions after labeling. However, results were obtained where human expertise won out, suggesting the importance of the diversity of expertise and contextual inference capabilities \cite{holmes2023evaluating}. Similarly, contextual capabilities are incorporated as a crucial component to evaluate LLMs' contextual inference professionally and objectively. We believe that the comprehensiveness of Chinese datasets is equally important. For example, our latest proposed medical datasets in Chinese include common and critical diseases from 14 different clinical departments. Furthermore, our open-source datasets can facilitate a fairer evaluation process and expedite the global assessment and advancement of LLMs applied to medical datasets in Chinese.

Many current models are data-hungry and necessitate labor-intensive labeling \cite{ghassemi2020review}. The advent of medical knowledge graphs and foundation models, which enable training without labeled data and professional medical knowledge, has driven the application of AI throughout the clinical workflow, including triage, diagnosis, and clinical management \cite{Levine2023.01.30.23285067,korngiebel2021considering,rao2023assessing}. Inspired by these advancements, we developed Dr. PJ, an LLM based on massive medical datasets in Chinese. Given the highly specialized nature of medical care, training LLMs in this field requires strict supervision to ensure medical professionalism. Simultaneously, humanistic care, a fundamental aspect of doctor-patient communication, is crucial for human-computer interaction \cite{10.1001/jama.2017.19198}. 
% Many current models are data-hungry and necessitate labor-intensive labeling \cite{ghassemi2020review}. The advent of medical knowledge graphs and foundation models, which enable the training without labeled data and professional medical knowledge, drives the application of AI in the whole clinical workflow, including triage, diagnosis, and clinical management \cite{Levine2023.01.30.23285067,korngiebel2021considering,rao2023assessing}. Inspired by these advancements, we develop Dr. PJ, an LLM based on massive Chinese medical data. Given the highly specialized nature of medical care, training large language models in this field requires strict supervision to ensure medical professionalism. Simultaneously, humanistic care, a fundamental aspect of doctor-patient communication, is crucial for human-computer interaction \cite{10.1001/jama.2017.19198}.
Unlike ChatGPT and ERNIE Bot, which are general AI models pretrained on general internet data, Dr. PJ was built for medical applications and has been trained using medical texts. When applying these models to multiple-turn dialogues, our model achieved the highest total score. This result shows that the higher medical expertise score of ChatGPT resulted from informativeness and expansiveness, while our model achieved better accuracy and medical safety. Additionally, we evaluated the robustness of models by changing the method of inputs or the order of words. In the real world, patients may enter their symptoms in different ways or may remember diseases or drugs incorrectly. The word order may also have an influence on natural language understanding \cite{pham2021order}. Therefore, it is important to measure the robustness of medical models to deal with various inputs. Dr. PJ had higher semantic consistency and lower complete error rate compared to ChatGPT, indicating better robustness. Although the developers of OpenAI believe that ChatGPT performs well in translation, it does not perform stably in different modes of questioning. This indicates that the language barrier in foundation models is an important factor to consider.

% =======================================
% table 7
% =======================================

% However, there are limitations in the evaluation system and LLMs development. Firstly, the evaluation criteria primarily rely on subjective scoring. Although this approach aligns with the characteristics of the medical profession, it can result in time and resource wastage. Secondly, the dataset scale is still confined. Thirdly, foundation models with a greater number of parameters have the potential to yield greater accuracy. It is plausible that we could enhance the training model's performance by utilizing more complex parameters. Future work will necessitate the combination of automated model evaluation and the enrichment of evaluation data by providing medical datasets to researchers for collaborative construction. Additionally, it should be noted that utilizing different prompts may have an impact on model output \cite{10.1145/3560815}. Consequently, future evaluations of prompts will also prove to be significant.

However, limitations remain in the evaluation system and LLMs development. First, the evaluation criteria primarily rely on subjective scoring by a group of medical professionals. Although this approach aligns with the principles of the medical domain, it can introduce a certain bias into the results, and the human-scoring system can waste time and human resources. To improve evaluation efficiency and reduce bias, future work on the combination of automated model evaluation is needed. Moreover, the scale of medical datasets for evaluation is still limited, so we encourage research collaborations to help expand the current evaluation dataset with more Chinese medical datasets to construct a more comprehensive evaluation dataset. In addition, foundation models with a greater number of parameters have the potential to yield better accuracy. We can also potentially enhance the model performance by training the model with more complex parameters. Finally, note that utilizing different prompts may have an impact on model output \cite{10.1145/3560815}. Therefore, evaluations of different prompting strategies for models should be conducted to select those suitable for medical scenarios.

%%%%%%%%%%%%%%%%%%%%%%%%%%%%%%%%%%%%%
%           5. Conclusion
%%%%%%%%%%%%%%%%%%%%%%%%%%%%%%%%%%%%%
\section{Conclusion}
\label{sec:conclusion}

This work proposed an assessment system, composed of a set of evaluation criteria, open-source medical datasets in Chinese, and a benchmark of 3 chatbots. 
% These criteria will be used with Chinese datasets of 34 cases prepared for open-source that includes medical multiple-turn dialogues and case reports, contributing to the growing body of research on chatbots and their potential use in the medical field. 
Medical experts evaluated the LLMs and found that 3 chatbots (ChatGPT, ERNIE Bot, and Dr. PJ) could understand patients' questions and provide logical answers. Through a comparison using the proposed evaluation criteria, we found that Dr. PJ outperformed the other 2 models with more accurate medical knowledge and humanistic care. 
% Developing specialized LLM such as Dr. PJ improve patient care and outcomes. 
Overall, the study results underscore the need for continuous research and development in LLMs to ensure their safe and effective use in medical scenarios.

\bibliography{anthology,custom}
\bibliographystyle{acl_natbib}

\appendix

\end{document}